\documentclass[letterpaper]{article} 
\usepackage{aaai25}  
\usepackage{times}  
\usepackage{helvet}  
\usepackage{courier}  
\usepackage[hyphens]{url}  
\usepackage{graphicx} 
\usepackage{natbib}  
\usepackage{caption} 
\usepackage{booktabs}
\usepackage{algorithm}
\usepackage{algorithmic}

\usepackage[dvipsnames]{xcolor}
\usepackage[most]{tcolorbox}
\usepackage{url}
\usepackage{threeparttable}

\usepackage{newfloat}
\usepackage{listings}
\DeclareCaptionStyle{ruled}{labelfont=normalfont,labelsep=colon,strut=off} 
\floatstyle{ruled}
\newfloat{listing}{tb}{lst}{}
\floatname{listing}{Listing}
\nocopyright

\title{Whose Side Are You On? Measuring Stance at Scale for Reddit Discourse Analysis During the Israel-Gaza War}

\author{
  Oded, Shahar\\
  Ben-Gurion University of the Negev\\
  \texttt{odedshah@post.bgu.ac.il}
}

\affiliations{

}

\begin{document}

\maketitle


\begin{abstract}
Large-scale discourse analysis often requires stance labels, yet for many high-salience political events such labels are unavailable at the comment level. We introduce and validate a comment-level stance measurement instrument that assigns Pro-Israel, Pro-Palestinian, or Undefined stance using Reddit comment text. The instrument is calibrated from a small, manually annotated gold-standard dataset and scaled through large language model-assisted labeling, achieving weighted-$F_{1}=0.88$ against held-out human annotations. To enable corpus-scale measurement, we operationalize the instrument as a lightweight classifier based on DistilBERT embeddings and an RBF SVM, which achieves weighted-$F_{1}=0.7935$ on the original gold-standard set. Applying this instrument, we measure the stance of approximately 2.7 million Reddit comments posted between 2023 and 2025 during the Israel-Gaza war. We demonstrate what this measurement enables by analyzing stance-conditioned asymmetries in sentiment, toxicity, and emotional framing across political affiliations. This work shows how validated, low-cost stance measurement can support large-scale descriptive analysis of polarized discourse, without relying on network structure or prior labeled corpora.
\end{abstract}

\section{Introduction}
\label{sec:introduction}

Understanding how polarized communities frame and re-frame conflicts in real time is central to computational social science: it lets scholars track opinion shifts, test theories of collective behavior, and inform moderation policy. Surveys or small, hand-annotated corpora give high internal validity but (i) capture only static snapshots, (ii) are costly to replicate, and (iii) suffer from self-selection and desirability biases.  Public social-media traces, by contrast, provide continuous, unsolicited evidence of how millions of users talk, react, and evolve---\textbf{if} stance can be measured reliably at scale.

\textbf{Knowledge gap.} Two routes dominate large-scale stance work. Closed-source LLM annotation is accurate yet incurs non-trivial monetary, environmental, and rate-limit costs and embeds opaque decision rules.  Lightweight classical models are cheaper but typically rely on bespoke linguistic features, user metadata, or network structure that rarely generalize. A transferable, low-cost measurement instrument for stance, requiring only raw text and a small seed of manual labels, remains missing.

\textbf{Why Reddit? Why this conflict?} Reddit hosts the largest open corpus on the 2023--2025 Israel-Gaza war (Kaggle lists $>$2 M comments) yet lacks stance labels.  The long-form, threaded style of the platform invites nuanced expression: sarcasm, coded language, implicitly signaled positions, making naive or zero-shot stance measurements brittle. As a result, rigorous, comment-level stance analysis of this conflict (or indeed of Reddit more broadly) is still scarce.

\textbf{Label-at-scale bottleneck.} Manual labeling is prohibitive beyond a few thousand comments, while high-volume LLM annotation faces prompt sensitivity, topic drift, and---at sustained rates---dollar-and-carbon costs.  For example, labeling 500k comments with GPT-4o mini at current research-tier prices already rivals a small survey budget, and API throttling blocks large-scale use.

\textbf{Our approach.} To enable stance measurement at scale, we use a three-phase process---\emph{seed-label $\rightarrow$ weak-supervise $\rightarrow$ self-train}---that balances annotation quality with scalability. The process (i) begins with a 1{,}057-comment gold-standard set, (ii) uses GPT-4o mini to generate weak labels for 38k additional comments (weighted-$F_{1}=0.88$ against held-out human annotations), and (iii) trains an RBF SVM on DistilBERT embeddings to produce a lightweight stance classifier, achieving weighted-$F_{1}=0.7935$ on the original gold set. While slightly below GPT-4o mini in absolute accuracy, this classifier is computationally inexpensive and suitable for corpus-scale deployment. We apply it to label the full 2.7M-comment dataset, surfacing approximately 850k polarized comments for downstream discourse analysis.

\vspace{0.5em}
\textbf{Contributions:}
\begin{itemize}
    \item \textbf{Measurement instrument:} A validated, comment-level stance measurement for polarized discourse, transferable across conflicts using raw text and minimal annotation.
    \item \textbf{Validation evidence:} Reliability and robustness evaluation against human annotations, demonstrating suitability for large-scale descriptive analysis.
    \item \textbf{Corpus scale measurement:} Stance labels for 2.7M Reddit comments on the Israel-Gaza war, enabling longitudinal and comparative discourse analysis.
    \item \textbf{Substantive findings:} Empirical analysis of stance-conditioned differences in sentiment, toxicity, and emotional framing.
\end{itemize}

\subsection{Illustrative Examples}

\begin{figure}[h]
\centering

\fcolorbox{MidnightBlue!70!black}{MidnightBlue!7}{%
  \parbox{0.96\linewidth}{%
    \textbf{\color{MidnightBlue!70!black}Pro-Israel}\par
    \vspace{2pt}
    \itshape ``Are you forgetting that Hamas is the one to blame for those?''%
  }%
}

\vspace{0.6em}

\fcolorbox{ForestGreen!70!black}{ForestGreen!7}{%
  \parbox{0.96\linewidth}{%
    \textbf{\color{ForestGreen!70!black}Pro-Palestinian}\par
    \vspace{2pt}
    \itshape ``The IOF used an ambulance and costumes to try to rescue him, and ended up getting themselves all killed in the effort. If he had just stayed a hostage, he would have most likely still been alive today.''%
  }%
}

\vspace{0.6em}

\fcolorbox{Gray!70!black}{Gray!7}{%
  \parbox{0.96\linewidth}{%
    \textbf{\color{Gray!80!black}Undefined}\par
    \vspace{2pt}
    \itshape ``Hang on. Who are the good guys?''%
  }%
}

\caption{Illustrative examples of polarized comments.}
\label{fig:polarized_quotes}
\end{figure}

In this study, stance refers to expressed alignment toward one of the primary conflict parties as manifested through linguistic framing, attribution of responsibility, moral positioning, or evaluative language. Stance is distinct from sentiment: comments may be negative in tone while still aligning with either side, and positive sentiment toward in-group actors does not necessarily imply hostility toward out-groups. Our construct therefore captures target-conditioned political alignment rather than general affect or topic mention. This operationalization allows stance to function as a measurement variable for downstream analysis of polarized discourse.

Section~\ref{sec:related-work} reviews prior stance-detection and weak-supervision research.  
Section~\ref{sec:methodology} details the dataset and pipeline.  
Sections~\ref{sec:results}--\ref{sec:discussion} present evaluation, substantive findings, limitations, and future work.


\section{Related Work}\label{sec:related-work}

Large-scale stance measurement intersects three strands of research: (i) \textit{stance classification models}, from classical learners to LLMs; (ii) \textit{low-resource annotation pipelines} that curb the cost of large labeled corpora; and (iii) \textit{computational analyses of online conflict}, with Reddit emerging as a rich but under-labeled venue. We review each in turn and situate our contribution.

\subsubsection{Stance Detection}

Stance detection refers to identifying whether a text expresses a favorable, opposing, or neutral view toward a target.  Twitter was the early test-bed, where partisan hashtags and mention networks supply weak labels; neural encoders such as BERT and XLNet later lifted accuracy but still required carefully curated gold sets. Imtiaz et al. (2022) \cite{imtiaz2022taking} reached $F_{1}=0.94$ on 979 tweets about the 2021 Israel-Palestine escalation, yet their model never generalized beyond that slice.  Ali et al. (2025) \cite{ali2025social} offer the first \emph{Reddit} stance benchmark for the same conflict (9,969 comments, three classes) and report 74\% macro-F1 with a Mixtral-8×7B prompt, but their corpus is two orders of magnitude smaller than ours and remains LLM-dependent.  Beyond the Middle East, Zhu et al. (2022) \cite{zhu2022reddit} released an 8M-comment Reddit dump of the 2022 Russo-Ukrainian war, noting a dominant pro-Ukraine framing yet providing no comment-level labels.

A complementary line looks at \emph{long-form} texts whose source identity is known.  Wei \& Santos (2020) \cite{wei2020narrative} trained BERT to predict whether news, policy documents, or textbook excerpts originated from Israeli or Palestinian outlets, achieving $F_{1}\!\approx\!0.91$.  Their results confirm transformers’ ability to capture subtle narrative framing, but the setting differs: labels derive from explicit source affiliation and documents are paragraphs to pages long, whereas social-media stance is shorter, more sarcastic, and often implicit.

Taken together, the literature still lacks (a) high-volume, comment-level stance labels on Reddit and (b) empirical evidence that lightweight classifiers, when bootstrapped through weak supervision, can approximate LLM-level labeling quality in such domains. Our study addresses both by evaluating this trade-off explicitly in a conflict setting where stance is subtle and often implicit.

\subsubsection{Low-resource and LLM-assisted annotation}
Early attempts to sidestep heavy manual labeling relied on \emph{distant supervision}---hashtags, subreddit membership, or graph propagation---but these heuristics drift as topics evolve and inject many false positives.  A second strand harnesses \emph{unsupervised / semi-supervised clustering}: for the 2018 Turkish elections, Rashed et al. (2021) \cite{rashed2021embeddings} grouped 1.8M Twitter users by Sentence-BERT embeddings, UMAP, and HDBSCAN, cleanly separating pro- and anti-Erdogan blocks without a single hand label.  Such approaches are inexpensive yet yield only \textit{user-level} clusters, limiting comment-level linguistic analysis.

Large language models now offer accurate zero-shot labels, but sustained use still faces rate limits, dollar cost, and opaqueness. Ali et al. (2025) \cite{ali2025social} estimate that even open-weight LLMs require substantial GPU resources for labeling 10k Reddit comments, and their best prompt misclassified neutral instances at non-trivial rates. Beyond financial and computational cost, reliance on proprietary LLM APIs raises concerns regarding reproducibility and measurement stability. Model versions evolve, prompt sensitivity shifts, and inference behavior may vary across time and deployment settings. For longitudinal discourse analysis, where stance functions as a measurement variable rather than a one-time annotation, stability and auditability are critical. Distilling an initial LLM-assisted phase into a lightweight classifier yields a fixed and inspectable decision boundary, enabling consistent measurement across time without dependence on evolving external systems.

In this study, we combine a brief human seed (1K comments) with LLM-assisted weak supervision (38K comments), and distill the resulting labels into a lightweight classifier suitable for repeated corpus-scale measurement. Importantly, we evaluate how closely such a classifier can approximate parallel LLM labeling in a domain requiring contextual and world knowledge. The resulting model operates at less than 0.25\$ per 1,000 comments and consumes under 0.15\% of the energy that a Mixtral-8×7B inference setup would require, assuming the near-linear relation between parameter count and inference energy documented by Strubell et al. (2020) \cite{strubell2020energy}. This second figure is a parameter-count extrapolation rather than a measured power draw, and should be read as an order-of-magnitude indication of the cost gap rather than a benchmark. While not state-of-the-art, this configuration demonstrates that reliable stance measurement at scale need not rely on sustained heavyweight LLM inference.

\subsubsection{Reddit and conflict discourse}

Reddit is no stranger to conflict research.  
Vachher \textit{et al.}~\cite{vachher2020understanding} (2020) modelled inter-community battles during \textit{r/place};  
Magu \textit{et al.}~\cite{magu2024understanding} (2024) traced echo chambers in U.S.\ gun-policy debates;  
and Zhu \textit{et al.}~\cite{zhu2022reddit} (2022) mapped stance asymmetries in the Russo-Ukrainian war.  
Yet none of these studies provides large-scale, \emph{comment-level} stance labels, so they cannot quantify toxicity or framing conditional on allegiance, nor do they provide validated, low-cost mechanisms for producing stance labels at scale.  
The only Reddit stance corpus for Israel-Palestine, released by Ali \textit{et al.}~\cite{ali2025social} (2025), covers only $\sim$10\,000 filtered comments and relies on heavyweight LLM inference.  
We complement this line of work by (i) releasing a 2.7 million-comment corpus with predicted labels, and (ii) showing that a lean SVM\,+\,DistilBERT classifier---bootstrapped from an LLM---reaches broadly comparable labeling quality at a small fraction of the inference cost, while retaining a fixed and inspectable decision boundary. This comparison is indicative rather than head-to-head: Ali et al. report macro-F\textsubscript{1} on their own $\sim$10k-comment split, whereas we report weighted-F\textsubscript{1} on a differently constructed gold set, so the two figures are not directly commensurable. A direct evaluation on their released split remains future work.

\subsubsection{Gap synthesis}

Across these strands, two blind spots persist: scalable, comment-level stance labeling on Reddit, and systematic behavioral analysis enabled by such labels in conflict settings. At the same time, stance in geopolitical conflicts often depends on contextual and world knowledge, making purely heuristic or distant-supervision approaches brittle. Our work addresses these gaps by demonstrating how minimal human annotation, LLM-assisted weak supervision, and lightweight classification can be combined to produce reliable stance labels at scale, without requiring sustained LLM inference. This configuration is portable to other conflict domains where labeled data are scarce and contextual knowledge is necessary for interpretation.


\section{Methodology}
\label{sec:methodology}

We designed a three-stage process---\emph{seed annotation}, \emph{LLM-assisted expansion}, and \emph{lightweight supervised modeling}---to support reliable stance labeling at the Reddit comment level for large-scale discourse analysis.

\subsection{Human gold set}
We begin with a human-labeled gold standard of 1{,}057 Reddit comments related to the Israel-Gaza war, sampled from the daily updated dataset available on Kaggle \cite{asaniczka_2024}, annotated by a trained annotator familiar with the conflict as Pro-Israel, Pro-Palestinian, or Undefined based on linguistic cues, framing, and tone. Comments expressing alignment with one side, including sarcastic or emotionally charged language, were labeled accordingly. This gold set serves both as a construct definition for stance and as a reference point for evaluating downstream labeling reliability. The implications of the single-annotator design are discussed in the Limitations (\S\ref{sec:limitations}).

\subsection{LLM‐assisted expansion}
To scale beyond this manually labeled core, we prompt-engineered GPT-4o mini to label an additional 38,000 conflict-related comments. Agreement with the human gold set ($\text{Accuracy}=0.8696$) was used to assess the suitability of the prompt for large-scale stance labeling, with iterative refinements applied to stabilize class boundaries on a development subset of 50--100 comments, later folded into training and excluded from the test set. We also added $\approx$5,000 unrelated Reddit comments (e.g., from a Russia--Ukraine discourse dataset \cite{ukraine_russia_reddit_data}) labeled as \textsc{Undefined}. Their purpose is deployment realism rather than class balance: the instrument is applied to an unfiltered corpus stream in which a substantial share of comments in conflict-related subreddits are off-topic, and such comments must not be forced into a polarized class. These out-of-domain instances make up roughly 16\% of the training \textsc{Undefined} pool, while the remaining $\approx$25.5k \textsc{Undefined} instances are in-domain Israel--Gaza comments carrying no discernible alignment, so within-topic neutrality remains the dominant signal for the class. We nonetheless flag in \S\ref{sec:limitations} that this construction risks entangling off-topic cues with the neutrality construct.

\begin{table}[t]
\centering
\caption{Class distribution in the human-labeled (gold) and GPT-labeled datasets.}
\resizebox{\columnwidth}{!}{%
\begin{tabular}{lcc}
\toprule
\textbf{Class} & \textbf{Gold Labels} & \textbf{LLM Labels} \\
\midrule
Pro-Israel      & 309   & 5,100  \\
Pro-Palestinian & 462   & 6,692  \\
Undefined       & 286   & 25,491 + 4,871 \\
\textbf{Total}  & \textbf{1,057} & \textbf{42,154} \\
\bottomrule
\end{tabular}}
\label{tab:samples_per_class_israel_palestine}
\end{table}

\subsection{Embedding and classifier}
Each comment was split into sentences; DistilBERT produced sentence embeddings that were mean-pooled to a comment vector. An RBF-kernel SVM---with class-weighted loss to offset the Undefined majority---served as the student classifier. Hyper-parameters were tuned via Optuna \cite{akiba2019optuna}; Alternative classifier heads (Logistic Regression, XGBoost, FFNN) and an end-to-end finetuned DistilBERT were also evaluated but did not offer comparable separation between polarized classes under this labeling setup.

\subsection{Evaluation}
We held out 1{,}057 human-labeled comments for final evaluation. All models were trained on the GPT-labeled corpus (80/20 train–validation split) and evaluated on this held-out set.

We report two F1-based metrics. First, the \emph{weighted F1 over all classes} (\(F1_{WT}\)), which computes per-class F1 and averages them using class frequencies:
\[
F1_{WT} \;=\; \sum_{c \in \mathcal{C}} w_c \cdot F1_c, 
\quad 
F1_c \;=\; \frac{2 \cdot \text{Prec}_c \cdot \text{Rec}_c}{\text{Prec}_c + \text{Rec}_c},
\]
where \(w_c = \tfrac{n_c}{\sum_{j} n_j}\) is the proportion of class \(c\) in the test set.

Second, the \emph{weighted F1 over polarized classes only} (\(F1_{WP}\)), defined analogously but with the class set restricted to \(\{\text{Pro-Israel}, \text{Pro-Palestinian}\}\) and the weights re-normalized:
\[
F1_{WP} \;=\; \sum_{c \in \{\text{Pro-Israel}, \text{Pro-Palestinian}\}} 
\tilde{w}_c \cdot F1_c.
\]

We use \(F1_{WP}\) to assess suitability for downstream discourse analysis because it emphasizes preserving separation between the two polarities rather than inflating performance by assigning ambiguous cases to \textsc{Undefined}. In short, we favor models with strong \(F1_{WP}\) and competitive \(F1_{WT}\), avoiding degenerate solutions that “run” to the \textsc{Undefined} class.

\subsection{Corpus-wide labeling and feature stack}
We applied the trained model to 2.7M Reddit comments (Oct.\ 2023--May 2025) and computed several features to analyze the rhetorical and thematic framing of political discourse at scale. Comparative results are presented in Section~\ref{sec:discussion}.

\begingroup\sloppy
\begin{itemize}
  \item \textbf{Toxicity}: \path{toxic-bert} \cite{toxicbert}.
  \item \textbf{Sentiment}: \path{twitter-roberta-base-sentiment} \cite{cardiffnlp_sentiment}.
  \item \textbf{Emotion}: \path{emotion-english-distilroberta-base} \cite{hartmann_emotion}. A comment's \emph{emotionality} is the total probability mass the model assigns to the six non-neutral Ekman categories (anger, disgust, fear, joy, sadness, surprise), equivalently $1-P(\textsc{neutral})$, from a single forward pass over the truncated comment. The score lies in $[0,1]$; higher values indicate greater emotional charge irrespective of valence, and group-level figures are means over comments.
  \item \textbf{Topics}: BERTopic \cite{grootendorst2022bertopic} on 80k stratified comments, clustered with Agglomerative + UMAP; super-topics named via GPT-4o mini (e.g., \textit{IDF}, \textit{HAMAS}).
\end{itemize}
\endgroup

\begin{figure}[h!]
\centering
\includegraphics[width=1\linewidth]{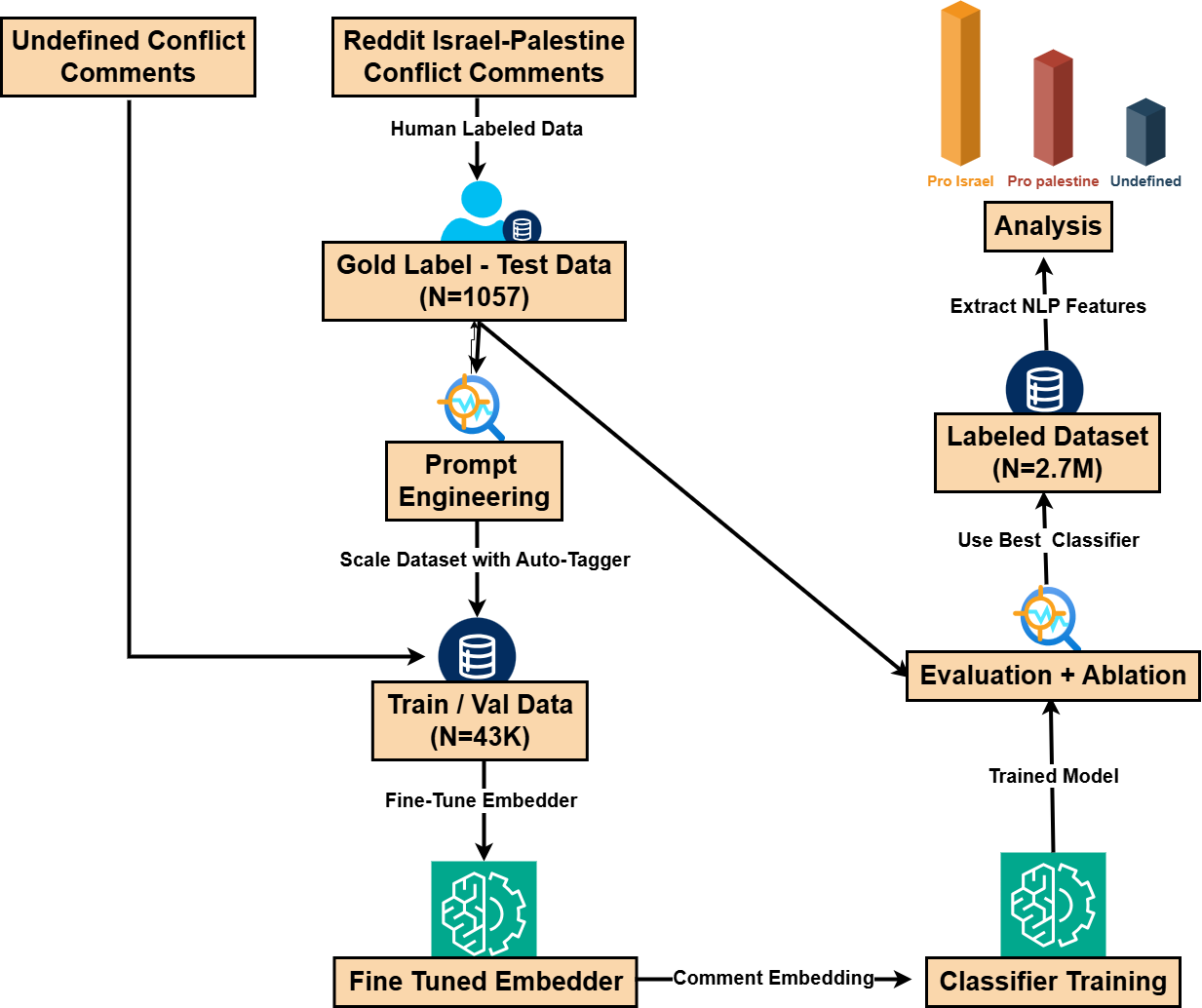}
\caption{Pipeline: human seed → LLM labels → DistilBERT embeddings → SVM student → corpus labeling and feature extraction.}
\label{fig:methodology-scheme}
\end{figure}

This process yields stance labels with reliability comparable to direct LLM annotation while remaining feasible for longitudinal measurement on large Reddit corpora. Full analysis appears in Section~\ref{sec:results}.

\subsection{External Validation Methodology}
To assess the external validity of the stance labeling measurement pipeline and its ability to generalize beyond the Israel--Gaza context to a distinct political domain with different actors and rhetorical conventions, we conducted an external validation on an independent political dataset. We utilized the Donald Trump subset of the P-Stance \cite{p_stance_data} corpus, which contains tweets originally labeled as either \textit{Favor} or \textit{Against}.

To align this dataset with our proposed three-way classification scheme, we introduced an \textsc{Undefined} category by sampling a random subset of comments from the Israel–Palestine Reddit \cite{asaniczka_2024} corpus. To ensure high-quality gold labels for evaluation, we manually re-annotated the entire test subset, resulting in three final classes: \textsc{Pro-Trump}, \textsc{Against-Trump}, and \textsc{Undefined}. This construction has a known limit: the \textsc{Undefined} instances come from a different platform and topic than the \textit{Favor}/\textit{Against} tweets, so the class does not test within-topic neutrality. In practice it did not make the task easier -- \textsc{Undefined} was by far the hardest class ($F_{1}=0.32$, against $0.72$ and $0.78$ for the two polarized classes) -- so the three-way figure is if anything depressed rather than inflated by it. The external result is therefore best read as evidence that the \emph{polarized} distinction transfers to a new political domain.

As in the primary task, prompt refinement used a development subset of 50--100 comments, later folded into training and excluded from the gold test subset. The process followed three stages: (1) zero-shot labeling to re-label the original P-Stance training and validation sets, intentionally disregarding their original labels; (2) downstream classifier training on this LLM-labeled data; and (3) evaluation against the manually annotated gold labels in the held-out test set.

\begin{table}[t]
\centering
\caption{Class distribution in the gold labels (test subset) and GPT-labeled external Trump datasets.}
\resizebox{\columnwidth}{!}{%
\begin{tabular}{lcc}
\toprule
\textbf{Class} & \textbf{Gold Labels} & \textbf{LLM Labels} \\
\midrule
Pro-Trump       & 295   & 2,464  \\
Against-Trump   & 415   & 3,846  \\
Undefined       & 67    & 869+1,033 \\
\textbf{Total}  & \textbf{777} & \textbf{8,212} \\
\bottomrule
\end{tabular}}
\label{tab:samples_per_class_trump}
\end{table}


\section{Results}\label{sec:results}

This section reports (i) validation of LLM-assisted labels and the resulting stance measure, and (ii) the discourse patterns enabled by applying this measure to 2.7M Reddit comments.

\subsection{LLM Labeling Reliability}

The LLM based labeling (Auto-Tagger) demonstrates high agreement with human annotations on Pro-Israel and Pro-Palestinian comments, with slight overclassification into the Undefined category. A high weighted F$_{1}=0.88$ demonstrates its reliability for distilling large-scale training data from unlabeled discourse, when prompted properly. To contextualize this result, we follow the evaluation setup of \cite{ali2025social}, which evaluated classical prompting methods (zero-shot, one-shot, few-shots) against an engineered instruction prompt. We compared the different prompt methods using our suggested base LLM (GPT-4o mini) for the task.

\begin{table}[t]
\centering
\caption{Auto-Tagger results on the gold test set.}
\resizebox{\columnwidth}{!}{%
\begin{tabular}{lcc}
\toprule
\textbf{Prompting Method} & \textbf{Weighted F$_{1}$} & \textbf{Accuracy} \\
\midrule
Zero-Shot           & 0.84   & 83.55\%  \\
One-Shot            & 0.81   & 79.96\%  \\
Few-Shots (3)       & 0.82   & 81.10\%  \\
\textbf{Engineered Prompt}  & \textbf{0.88} & \textbf{86.96\%} \\
\bottomrule
\end{tabular}}
\label{tab:autotagger-results}
\end{table}

On a shared test set ($N=1057$), the engineered prompt (86.96\% accuracy) compared to the zero-shot prompt (83.55\%) shows a statistically significant improvement by McNemar’s test (paired, two-sided), $p \leq 0.047$. The improvement from the reconstructed benchmark experiment possibly stems from the change of LLM (GPT-4o mini versus Mixtral-8$\times$7B), as well as the use of a different test set and tagging procedure.

The Auto-Tagger performed well on the external dataset as well, reaching $91.89\%$ accuracy and a weighted F$_{1}=0.92$ labeling the Trump discourse comments with our augmentation compared to the gold labels. This result further supports the suitability of LLMs as annotators for political stance in domains requiring contextual knowledge.

\subsection{Classifier Validation}

We evaluated multiple classifiers using two feature representations: DistilBERT embeddings and TF-IDF vectors, and compared the results to a full DistilBERT classifier, fine-tuned for the classification task. Classifier evaluation appears in Table~\ref{tab:final-results}. Because our goal is reliable measurement rather than architectural optimization, we treat model comparison as an adequacy check rather than a search for state-of-the-art performance. Recall that \(F1_{WT}\) is computed over all three classes, whereas \(F1_{WP}\) is restricted to the two polarized classes and therefore reflects only how well the instrument separates Pro-Israel from Pro-Palestinian. A model can raise \(F1_{WT}\) simply by routing ambiguous comments to \textsc{Undefined}, which is precisely the failure mode \(F1_{WP}\) is designed to expose.

\noindent\textbf{Model choice.} Among the evaluated options, the SVM offers the most favorable trade-off for preserving separation between polarized classes (\(F1_{WP}=0.8192\)) while keeping overall quality competitive (\(F1_{WT}=0.7935\)). It narrowly trails the fully fine-tuned DistilBERT model in \(F1_{WT}\) (0.8003) but better preserves separation between the two polarized classes, rather than inflating accuracy by routing ambiguous cases to \textsc{Undefined}. Logistic Regression and XGBoost on the same embeddings are close but consistently lower, and all TF-IDF variants perform substantially worse on both metrics (\(F1_{WT}\le 0.46\)). Within this measurement setting, representation quality appears to matter more than classifier depth. 

Given the fairness-sensitive objective and the need for stable separation between polarized classes, we select \emph{DistilBERT + SVM} for downstream analyses.

\begin{table}[h!]
\centering
\caption{Classifier performance on the held-out human gold test set ($N=1{,}057$).}
\label{tab:final-results}
\resizebox{\columnwidth}{!}{%
\begin{tabular}{lcc}
\toprule
\textbf{Model} & \textbf{\(F1_{WT}\)} & \textbf{\(F1_{WP}\)} \\
\midrule
\textbf{DistilBERT (full clf)}     & \textbf{0.8003} & 0.8030 \\
\textbf{DistilBERT + SVM} & 0.7935 & \textbf{0.8192} \\
DistilBERT + XGBoost      & 0.7736 & 0.7939 \\
DistilBERT + LogReg       & 0.7865 & 0.8066 \\
DistilBERT + FFNN         & 0.7756 & 0.7999 \\
TF-IDF + SVM              & 0.4568 & 0.3773 \\
TF-IDF + XGBoost          & 0.3893 & 0.4428 \\
TF-IDF + LogReg           & 0.3565 & 0.3907 \\
TF-IDF + FFNN              & 0.4578 & 0.3773 \\
\bottomrule
\end{tabular}}
\end{table}

Because stance labels are subsequently used as analytical variables, it is important to characterize the nature of residual classification error.

\paragraph{Misclassification Profile.}
Inspection of the confusion matrix on the gold validation set reveals limited but non-negligible cross-polar misclassification. Specifically, 40 Pro-Israel comments were predicted as Pro-Palestinian (12.9\% of the 309 true Pro-Israel cases), and 30 Pro-Palestinian comments were predicted as Pro-Israel (6.5\% of the 462 true Pro-Palestinian cases). Cross-polar error is thus asymmetric: the instrument is roughly twice as likely to pull a Pro-Israel comment across the boundary as the reverse. Such cross-polar errors attenuate between-group differences in downstream analyses by mixing group means toward one another. These empirically estimated confusion rates are therefore used to implement measurement-error corrections in subsequent discourse-level statistical tests.

\paragraph{External Validation.}
Similarly, a DistilBERT-based classifier fine-tuned using our proposed methodology on the augmented \textit{P-Stance} Trump dataset \cite{p_stance_data} achieved an $F1_{WT}$ score of $0.72$ and an $F1_{WP}$ score of $0.76$. This comparable performance on an external dataset supports the robustness of the stance labeling process across political domains.

We also tested a directly fine-tuned DistilBERT classifier trained only on the 1K human-labeled examples, without access to the GPT-augmented data, on a stratified 80/10/10 train--validation--test split of that set. The resulting test partition holds only 106 comments, so this figure carries a correspondingly wide confidence interval and is reported as an indicative baseline rather than a precise estimate. 

\begin{table}[h!]
\centering
\caption{Fine-Tuned Classifier performance on held-out test set.}
\label{tab:ft-comparison-results}
\resizebox{\columnwidth}{!}{%
\begin{tabular}{lcc}
\toprule
\textbf{Model} & \textbf{\(F1_{WT}\)} & \textbf{\(F1_{WP}\)} \\
\midrule
DistilBERT (full clf)     & 0.719 & 0.713 \\
\bottomrule
\end{tabular}}
\end{table}

The significant drop in performance from the DistilBERT classifier trained on the augmented labels highlights the practical value of high-quality, auto-generated labels when large manually labeled datasets are unavailable.

\subsection{Stance Prevalence and Participation Dynamics}

Once the final classification method (SVM + DistilBERT embeddings) was selected, it was applied to the full dataset of 2.7M Reddit comments spanning October 2023 to May 2025. Table~\ref{tab:label-distribution} reports the aggregate distribution of predicted stance labels, indicating that approximately 31\% of comments express a polarized position, while the remainder are classified as \textsc{Undefined}.

\begin{table}[h!]
\centering
\caption{Predicted stance breakdown for 2.7M comments ($\approx 850$k polarized).}
\label{tab:label-distribution}
\resizebox{\columnwidth}{!}{%
\begin{tabular}{lrr}
\toprule
\textbf{Label} & \textbf{Count} & \textbf{Proportion} \\
\midrule
Pro-Israel         & 440,587 & 16.33\% \\
Pro-Palestinian    & 407,135 & 15.09\% \\
Undefined          & 1,850,208 & 68.57\% \\
\textbf{Total}     & \textbf{2,697,930} & \textbf{100\%} \\
\bottomrule
\end{tabular}}
\end{table}

\begin{figure}[tb] \centering \includegraphics[width=1.05\linewidth]{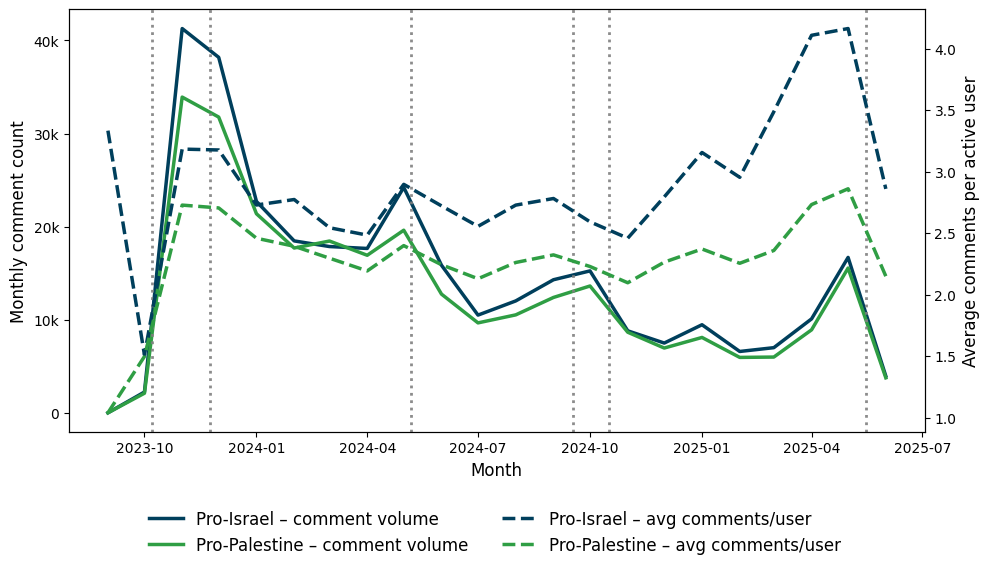} \caption{Monthly polarized comment volume (solid lines, left axis) and average comments per active user (dashed lines, right axis) by stance group. Vertical dotted lines indicate major conflict-related events (e.g. Oct 7th attack, first ceasefire, Rafah operation, beeper operation etc.).} \label{fig:stance-trend} \end{figure}

Figure~\ref{fig:stance-trend} presents two complementary indicators of stance participation over time: monthly polarized comment volume (solid lines, left axis) and average comments per active user (dashed lines, right axis).

At the level of aggregate volume, two descriptive patterns are evident. First, Pro-Israel and Pro-Palestinian participation move in close parallel across the 20-month window: month-to-month leads alternate and neither stance persistently dominates overall activity on Reddit. Second, participation is event-responsive. Following the initial surge in October–December 2023, overall activity declines but spikes recur around salient conflict events.

Volume parity, however, does not imply symmetric participation structure. When conditioning on active users, clear differences emerge. Around late 2024 and early 2025, including the period surrounding the \emph{Merkavot Gideon} operation, Pro-Israel participation is characterized by elevated per-user activity, whereas Pro-Palestinian intensity remains comparatively stable. In these periods, similar aggregate volumes arise from distinct engagement profiles: one driven by higher output per participant, the other by steadier contribution rates. 

This divergence suggests asymmetric engagement intensity rather than purely symmetric collective response, illustrating how stance-aware measurement reveals participation dynamics that aggregate counts alone would obscure.

\noindent\emph{Scope note.} These findings describe participation behavior within Reddit and should not be interpreted as population-level support or public opinion. Observed patterns are contingent on platform demographics, sub-reddit composition, moderation norms, and classifier error.

\subsection{Discourse Asymmetries in Polarized Participation}

Having established stance prevalence and participation dynamics, we now examine how polarized groups differ in rhetorical and affective expression. Rather than focusing solely on participation volume, we analyze community reception, affective polarity, emotional intensity, and toxicity to characterize stance-conditioned discourse asymmetries at scale.

Group comparisons are conducted using Welch's $t$-tests to account for unequal variances. Unless stated otherwise, a group mean is the unweighted mean over all comments assigned that stance; topic-level means are taken over the comments assigned to the topic in question. Topic-level claims are restricted to topics carrying at least 1{,}000 comments in \emph{each} stance group, which excludes four sparse domains (for example \emph{Two-State Solution}, $n=326$) whose group estimates are too unstable to interpret. Figures plot all topics for completeness; the threshold governs only which topics we draw claims from. Because stance labels are inferred rather than directly observed ($F1_{WT}=0.7935$), observed group means are subject to misclassification bias. Let $\boldsymbol{\mu}^{obs}$ denote the vector of observed group means and $P$ the confusion matrix restricted to the two polarized classes, row-normalized over polarized predictions, and estimated on the gold validation set. We obtain corrected estimates via

\[
\boldsymbol{\mu}^{true} = P^{-1} \boldsymbol{\mu}^{obs}.
\]

Under non-differential misclassification, such errors attenuate between-group differences toward zero. This correction assumes stable error rates estimated on the validation set and is intended to reduce attenuation bias rather than to increase effect sizes. We therefore report both observed ($p_{obs}$) and confusion-matrix–corrected ($p_{corr}$) test statistics; corrected estimates should be interpreted as robustness checks rather than inflated effects. In practice, corrections generally increase between-group gaps, consistent with attenuation under cross-polar misclassification.

Two caveats govern how these corrected values should be read. First, inverting $P$ amplifies estimation noise in its off-diagonal entries, so $\boldsymbol{\mu}^{true}$ is necessarily estimated less precisely than $\boldsymbol{\mu}^{obs}$. Second, the correction presumes \emph{non-differential} misclassification---that classifier error is independent of the outcome under comparison---which need not hold if, for instance, highly toxic or heavily sarcastic comments are also systematically harder to classify. We therefore treat the corrected gap as an upper bound and the observed gap as a lower bound on the plausible effect; every substantive conclusion reported below holds at both ends of that interval, and none rests on the corrected value alone.

\paragraph{Community Reception.}
We first assess how stance-conditioned comments are received by the Reddit community. Pro-Israel comments receive significantly higher average net scores (upvotes minus downvotes) than Pro-Palestinian comments. This difference suggests asymmetric endorsement patterns: stance expression is not only produced differently but also evaluated differently by other users on the platform.

\begin{table}[t]
\centering
\begin{threeparttable}
\caption{Average net comment score by stance group.}
\label{tab:engagement}
\begin{tabular}{l c p{3cm}}
\toprule
\textbf{Stance Group} & \textbf{Avg. Score} & \textbf{Interpretation} \\
\midrule
Pro-Israel        & 20.03 & Higher community approval \\
Pro-Palestinian   & 14.77 & Lower endorsement \\
\bottomrule
\end{tabular}
\begin{tablenotes}[flushleft]\footnotesize
\item Observed difference: $\Delta_{obs}=5.26$ (Welch’s $t$-test, $p_{obs}<10^{-4}$). Confusion-matrix–corrected difference: $\Delta_{corr}=6.78$ ($p_{corr}<10^{-4}$).
\end{tablenotes}
\end{threeparttable}
\end{table}

\paragraph{Affective Polarity.}
Average sentiment polarity is negative for both stance groups throughout the observation window, consistent with the conflict-centered nature of the discourse. While Pro-Palestinian comments are slightly more negative on average than Pro-Israel comments, the magnitude of this difference is substantively small (Cohen’s $d \approx 0.05$). 

Correcting for cross-polar misclassification increases the estimated gap, but the overall difference remains modest relative to within-group variation. Temporal dynamics are also broadly parallel: both groups exhibit similar fluctuations over time, including partial moderation around major events such as the November 2023 ceasefire.

Importantly, the most pronounced sentiment differences emerge at the topical level rather than in aggregate. Identity-salient domains, particularly \emph{Palestinian Identity}, show larger polarity gaps than policy-oriented discussions. Generalized negativity therefore cannot explain the observed asymmetries in engagement or aggression.

\begin{figure}[tb]
\centering
\includegraphics[width=1.05\linewidth]{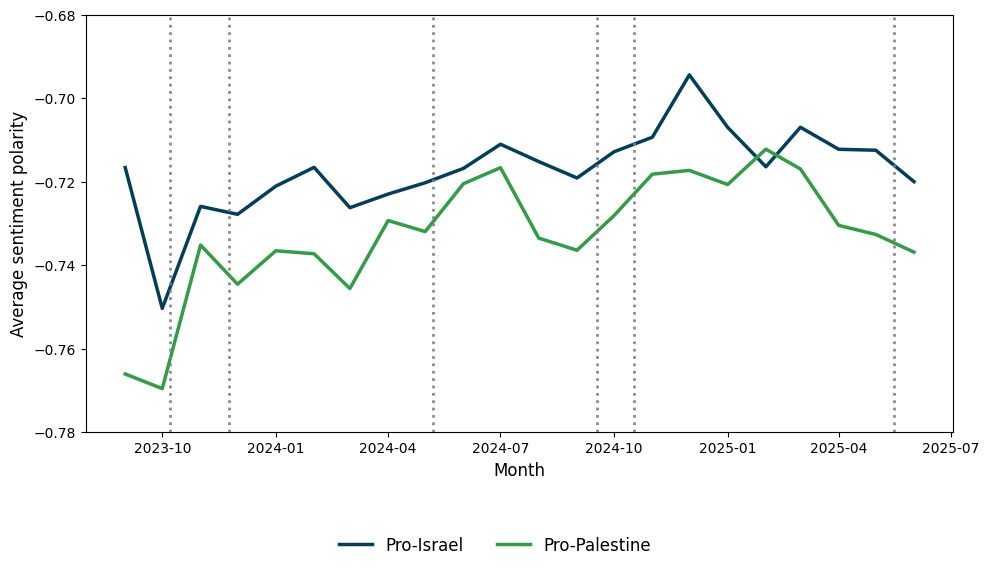}
\caption{Monthly average sentiment polarity for polarized comments. Both stance groups exhibit persistently negative and largely parallel affective trajectories.}
\label{fig:sentiment-trend}
\end{figure}

\paragraph{Aggression and Norm Violation.}
We next examine toxicity as a proxy for norm-violating or aggressive language. At the aggregate level, Pro-Palestinian comments exhibit higher average toxicity scores than Pro-Israel comments. This difference remains statistically robust after correcting for cross-polar misclassification, and the correction increases the estimated gap, indicating attenuation in the uncorrected comparison.

Importantly, toxicity differences are strongly topic-contingent rather than uniform. The largest corrected gaps emerge in identity-salient and morally charged domains, including \emph{Palestinian Identity}, \emph{Zionism}, \emph{Antisemitism}, and \emph{Settlements/Ethnic Cleansing}. By contrast, procedural and meta-level discussions such as \emph{Ceasefire Talks} and \emph{Media \& Propaganda} show markedly smaller differentials.

These patterns suggest that norm-violating language appears selectively amplified in contexts tied to collective identity and moral grievance, rather than evenly distributed across all areas of discourse.

\begin{figure}[tb]
\centering
\includegraphics[width=1.0\linewidth]{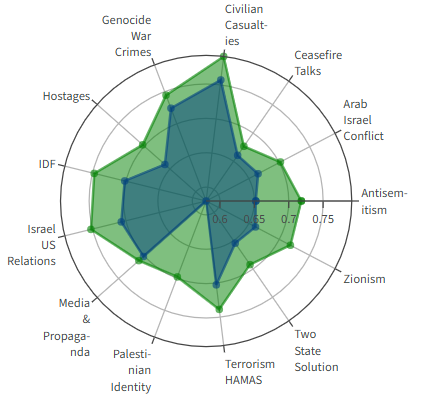}
\caption{Average toxicity scores by topic for Pro-Israel and Pro-Palestinian comments.
\emph{Note:} Overall means (TOTAL): Pro-Israel $=0.6677$, Pro-Palestine $=0.7090$. Observed gap (PI$-$PP) $\Delta_{\text{obs}}=-0.0413$; confusion-corrected gap $\Delta_{\text{corr}}=-0.0532$. Welch’s $t$-test: $p_{\text{obs}}<10^{-4}$, $p_{\text{corr}}<10^{-4}$.}
\label{fig:toxicity-topic}
\end{figure}

\paragraph{Emotional Intensity.}
While sentiment polarity differs only modestly between stance groups, emotional intensity shows a clearer asymmetry. Pro-Palestinian comments exhibit higher average emotionality scores overall, and this difference strengthens after correcting for cross-polar misclassification. 

Unlike sentiment, emotionality gaps are not diffuse but cluster in specific topical domains. The largest corrected differences appear in discussions of \emph{IDF}, \emph{Settlements/Ethnic Cleansing}, \emph{Civilian Casualties}, \emph{Hostages}, \emph{Terrorism/HAMAS}, and \emph{Palestinian Identity}. These are contexts in which moral evaluation, perceived injustice, and collective identity are especially salient. The pattern is not universal: \emph{Media \& Propaganda} reverses it, with Pro-Israel comments scoring higher on emotionality, which reinforces that the asymmetry is topic-contingent rather than a global property of either group.

Taken together, these findings indicate that stance-conditioned differences manifest more strongly in expressive intensity than in evaluative polarity. The asymmetry is therefore not primarily about how negative discourse is, but about how emotionally charged it becomes in identity-linked contexts. This pattern is consistent with research on moralized political identity and moral outrage, which shows that identity-linked grievances are often communicated through heightened affect and boundary-challenging language rather than simple shifts in valence \cite{skitka2014social,brady2017emotion}.

\begin{figure}[tb]
\centering
\includegraphics[width=1.0\linewidth]{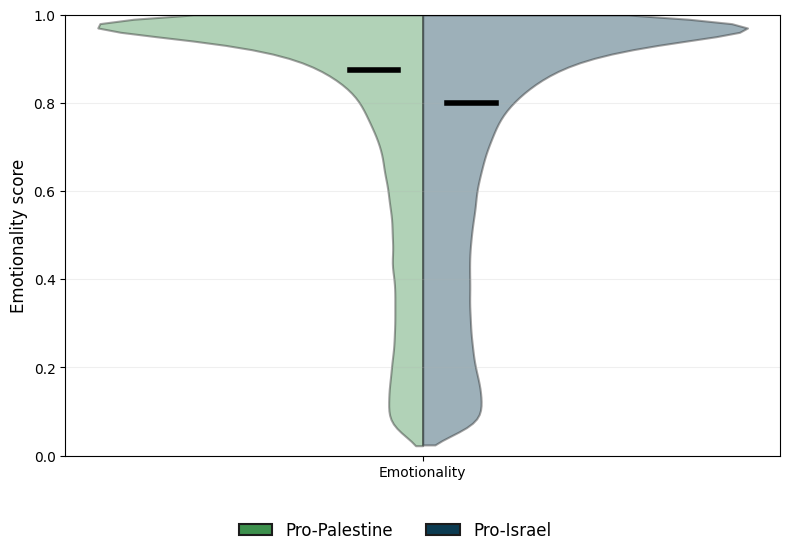}
\caption{Split violin plot of emotionality scores by stance (left: Pro-Palestine; right: Pro-Israel). Shapes represent kernel densities; black bars denote medians. Pro-Palestinian comments exhibit a distributional shift toward higher emotional intensity and a heavier upper tail. 
\emph{Note:} Means (TOTAL): PI $=0.6853$, PP $=0.7396$; $\Delta_{\text{obs}}=-0.0543$, $\Delta_{\text{corr}}=-0.0700$; Welch’s $t$: $p_{\text{obs}}, p_{\text{corr}}<10^{-4}$.}
\label{fig:emotionality}
\end{figure}

\paragraph{Synthesis.}
Taken together, these findings reveal a structured asymmetry in polarized discourse. While both stance groups exhibit comparably negative sentiment, differences emerge in affective intensity, norm-violating language, and community reception. Pro-Israel comments receive higher net endorsement, whereas Pro-Palestinian comments display elevated toxicity and emotional intensity, particularly in identity-salient domains.

This configuration is descriptively inconsistent with a simple negativity asymmetry. Instead, it reflects divergence in expressive modality: one pattern characterized by greater community approval and comparatively moderated tone, the other by intensified affect and greater use of language that crosses conventional discourse boundaries. 

Such patterns are consistent with theories of moralized political identity and moral outrage, which posit that identity-linked grievances are communicated through heightened emotional activation and norm-boundary transgression, rather than through shifts in valence alone. Importantly, these asymmetries remain visible after correcting for cross-polar misclassification, suggesting they are not artifacts of automated labeling but structural features of stance-conditioned discourse within this platform context.


\section{Discussion \& Future Work}\label{sec:discussion}

This study demonstrates that a small, carefully curated human seed (1K comments), amplified via GPT-4o mini and distilled into contextual embeddings with an SVM classifier head, can reliably label 2.7M Reddit comments with weighted-$F_{1}=0.7935$ at a fraction of the monetary and computational cost of large-scale LLM inference. Beyond efficiency, the resulting labels enable substantive analysis of stance-conditioned discourse asymmetries at scale.

\subsection{Contributions}

\paragraph{Measurement contribution.}
We introduce a lean, hybrid labeling pipeline that combines limited human annotation, LLM-assisted expansion, semantic embeddings, and classical classifiers. Unlike end-to-end LLM tagging, this approach is computationally inexpensive and transparent. Crucially, we explicitly model and correct for misclassification bias using the inverse confusion matrix, reducing attenuation in downstream statistical tests. This provides a replicable template for stance inference under realistic annotation constraints.

\paragraph{Substantive contribution.}
Empirically, we find that polarized discourse asymmetries on Reddit are not primarily characterized by differences in overall negativity. Instead, stance groups diverge in expressive intensity, toxicity, and community reception. Pro-Israel comments receive higher net endorsement, whereas Pro-Palestinian comments exhibit elevated emotional intensity and toxicity, particularly within identity-salient domains. These findings suggest that expressive asymmetry manifests more through affective activation and norm-boundary crossing than through shifts in evaluative polarity alone.

\paragraph{Computational efficiency contribution.}
Our results also highlight diminishing returns from deeper classifier heads under limited gold data. Contextual embeddings account for most of the predictive gain, whereas more complex architectures add marginal improvement. This supports the use of lightweight classifiers for large-scale monitoring tasks, reducing both financial and environmental costs relative to direct LLM inference.

\subsection{Limitations}\label{sec:limitations}

Despite encouraging results, several limitations remain:

\begin{itemize}
\item \textbf{LLM-generated training labels.} Although LLM labels align with human annotations ($\text{Accuracy}=0.8696$), they may embed systematic biases and under-represent rare constructions. Prior work reports political bias in LLM outputs \cite{ali2025social}; if present, such bias could propagate systematically through the weak-supervision stage and influence class boundaries at scale.

\item \textbf{Subjectivity of stance attribution.} Stance labeling is inherently interpretive, especially in cases of sarcasm, irony, implicit criticism, or mixed positioning. Even trained annotators may disagree in borderline cases. In addition, stance labeling was performed by a single trained annotator. While this ensured construct consistency during instrument calibration, it does not permit inter-annotator reliability estimation and may embed individual interpretive bias. Consequently, every agreement figure reported here---for the LLM labeler and for the student classifier alike---measures consistency with a single rater rather than against an adjudicated consensus, and should be read accordingly. Establishing inter-annotator reliability on a substantial subset is the most direct way to strengthen the instrument and is the first item of planned future work.

\item \textbf{Stability of misclassification rates.} The inverse confusion-matrix correction assumes that misclassification rates estimated on the gold validation set remain approximately stable over time. Shifts in discourse style, rhetoric, or topic salience across the observation window may violate this assumption and affect corrected estimates.

\item \textbf{Construction of the \textsc{Undefined} class.} Roughly 16\% of training \textsc{Undefined} instances are drawn from an unrelated conflict corpus. This improves robustness to off-topic input at deployment, but risks the classifier partly learning \emph{topical} off-domain cues as a proxy for neutrality rather than within-topic neutrality alone. Curating and human-adjudicating in-domain ambiguous comments would isolate the construct more cleanly.

\item \textbf{Platform specificity.} Findings reflect Reddit’s norms, moderation policies, and user demographics. They should not be interpreted as population-level attitudes or offline behavioral tendencies.

\item \textbf{Limited conversational context.} Comments are classified in isolation. Incorporating thread structure, temporal context, or user history may reduce ambiguity and improve stance inference.

\item \textbf{Proxy measurement of affective constructs.} Toxicity and emotionality scores are derived from pretrained models and function as proxies rather than validated psychological measures. Domain shift and cultural bias may differentially affect groups.
\end{itemize}

\subsection{Future Directions}

Several extensions follow naturally:

\begin{itemize}
\item \textbf{Cross-platform validation.} Applying the pipeline to other platforms or multilingual corpora would clarify whether expressive asymmetries generalize beyond Reddit.

\item \textbf{Context-aware stance modeling.} Integrating conversational structure and user-level features may improve disambiguation in reactive exchanges.

\item \textbf{Fine-tuning and domain adaptation.} While lightweight classifiers proved sufficient here, supervised fine-tuning of compact transformer models on curated stance corpora may further improve recall for rare rhetorical forms.

\item \textbf{Rhetorical and framing analysis.} Extending the pipeline to detect narrative strategies such as moralization, victimization, or whataboutism would enable deeper theoretical integration with political communication research.

\item \textbf{Longitudinal polarization dynamics.} Combining stance inference with network structure and temporal modeling could illuminate how expressive asymmetries evolve during critical political events.
\end{itemize}

Taken together, the findings illustrate how transparent, resource-efficient modeling can support fine-grained analysis of polarized discourse while maintaining methodological caution regarding inference scope.


\section*{Ethics and Data Statement}\label{sec:ethics}

This study analyzes only publicly available, previously collected Reddit and Twitter data distributed through public research datasets \cite{asaniczka_2024,ukraine_russia_reddit_data,p_stance_data}. We collected no data ourselves, contacted no accounts, and performed no intervention of any kind. As the work involves no interaction with human participants and uses only previously published public data, it was not submitted for institutional review board approval.

Reporting is deliberately aggregate. All analyses are presented at the level of stance group, month, and topic; we neither construct nor report user-level stance profiles, because comment-level predictions aggregated per account could be repurposed for targeting or harassment of individuals during an active conflict. We release code, prompts, and evaluation scripts, together with our stance predictions over a subset of an already-public corpus \cite{asaniczka_2024}. We redistribute no data of our own collection and introduce no identifying information beyond what that corpus already publishes. Illustrative quotations in Figure~\ref{fig:polarized_quotes} are short excerpts.

Two interpretive boundaries warrant emphasis. First, the instrument measures the stance \emph{expressed by a comment}, not the beliefs of its author: a single comment is a weak signal about a person, and the instrument is not fit for individual-level inference. Second, the descriptive asymmetries reported in Section~\ref{sec:results} characterize discourse on one platform across one time window and carry classifier error; they are not evidence about either national community, and they bear on how positions are expressed online rather than on the substantive merits of the conflict.


\section*{Code Availability}\label{sec:code-availability}

All code used in this study is publicly available on GitHub:

\begin{itemize} 
    \item \textbf{Israel-Palestine Political Affiliation Text Classification:} \url{https://github.com/shaharoded/Israel-Palestine-Political-Affiliation-Text-Classification}, containing the full stance classification pipeline, including automated labeling, contextual embedding generation, model training, and evaluation scripts. 
    
    \item \textbf{Israel-Palestine War Reddit Analysis:} \url{https://github.com/shaharoded/Israel-Palestine-War-Reddit-Analysis}, providing feature extraction modules (sentiment, toxicity, emotionality), statistical analysis scripts, and visualization code used to generate the figures in this paper.
\end{itemize}

Repositories include preprocessing scripts and modular components to facilitate replication and adaptation to other datasets.

An interactive dashboard presenting the full analysis is available at:
\url{https://israel-palestine-war-reddit-comments-analysis.streamlit.app/}.

\bibliography{references}


\end{document}